\documentclass[10pt,letterpaper,twocolumn]{article}

\usepackage[T1]{fontenc}
\usepackage[utf8]{inputenc}
\usepackage{lmodern}
\usepackage[letterpaper,margin=0.75in]{geometry}
\usepackage{graphicx}
\usepackage{cite}
\usepackage{xcolor}
\usepackage{microtype}
\usepackage{hyperref}
\usepackage{titlesec}
\usepackage{flushend}

\definecolor{linkblue}{HTML}{1F5A84}
\hypersetup{
  colorlinks=true,
  linkcolor=linkblue,
  citecolor=linkblue,
  urlcolor=linkblue,
  pdftitle={From Prompt to Harness: Coderlet from Scratch},
  pdfauthor={Mengfan Li},
  pdfsubject={A compact description of request flow, harness boundaries, state continuity, and bootstrapping}
}

\titleformat{\subsection}[hang]
  {\normalfont\normalsize\bfseries\raggedright}
  {\thesubsection}{1em}{}

\title{From Prompt to Harness: Coderlet from Scratch}
\author{Mengfan Li\thanks{Project repository: \url{https://github.com/lilinxi/Coderlet}}}
\date{}

\begin{document}

\maketitle

\begin{abstract}
A model alone does not determine how a programming agent acts. What the model sees, how actions enter the environment, how feedback returns, and how one run affects the next all depend on how the harness is organized. Minimal examples usually show only the basic interaction between a model and tools, while production systems spread these relationships across complex components and dependencies. This paper studies a compact harness design by following a single request through context formation, model decision, environmental action, observation return, and state continuation. Three boundaries---model, execution, and state---connect the model service, tool environment, and persistent state, while the request lifecycle determines the order in which these transitions occur. Together, they show the harness's core role: turning model generations into environmental actions, carrying runtime feedback into later decisions, and allowing state to continue across requests. On top of this runtime structure, a harness can also be gradually refined across runs through continued bootstrapping. The design is realized in the executable artifact \href{https://github.com/lilinxi/Coderlet}{\textbf{Coderlet}}.
\end{abstract}

\section{Introduction}

Model output takes several forms---text, code, and structured tool calls---but it becomes agent behavior through a separate control layer. The harness assembles what the model receives, evaluates proposed calls, dispatches accepted actions, turns tool results into observations, and decides what enters later history. In doing so, it turns isolated responses into a continuing interaction with an environment.

Small demonstrations make this interaction easy to see, although most reduce it to a loop between a model and one or two tools. Production systems add configuration, context limits, tool schemas, persistence, progress reporting, retries, and external dependencies, often burying the underlying flow. We keep that flow explicit and treat the surrounding components as extensions of it.

One request passes through five observable points. The model proposes content or an action. The harness decides whether to run the action, the environment returns a tool result, and the harness chooses what enters the request record. Progress updates reach the caller along the way. A successful request ends with a completion notification. Each point leaves different evidence: rejection still returns an observation without changing the environment, and an interrupted completion notification leaves an already saved request record intact.

The running example is a compact Codex-like local programming agent. Here, ``from scratch'' refers to building the control layer that prepares context, calls a model service, handles tools, stores history, and sustains the interaction. The model service and the OpenAI Codex product remain outside this scope. Limiting the example to the control layer keeps the request lifecycle visible. This is an implementation boundary rather than a claim of equivalent features or performance.

Following a request from input to its request record makes the model, execution, and state boundaries visible. Within this design, the read--decide--act--observe loop also offers a route for inspecting and improving mechanisms used in later runs. Our focus is the implementation and its request flow, not a complete production architecture.

The discussion moves from prior agent work to the request runtime and its boundaries, then closes with bootstrapping across runs.

\section{Related Work}

\subsection{Interaction and Memory}

Work on interactive agents shares a concern with how external information enters later decisions. ReAct, Toolformer, and SWE-agent study reasoning--action cycles, model-issued API calls, and agent--computer interfaces \cite{react,toolformer,sweagent}. Reflexion and Voyager study feedback retained for later decisions \cite{reflexion,voyager}. Their different mechanisms leave runtime coordination between model responses as a separate design question. We examine that layer---context preparation, tool dispatch, observation return, and state retention---rather than proposing a new reasoning or memory method.

\subsection{Outcomes and Traces}

Evaluation work separates task outcomes from process evidence. SWE-bench measures task outcomes, AgentBoard exposes intermediate progress, and harness-aware evaluation treats the surrounding runtime as an important experimental choice \cite{swebench,agentboard,vatsgolev}. These views complement one another, yet an outcome score and a trace support different claims. We use that distinction to separate proposed calls, executed tool actions, returned observations, and request records.

\subsection{Harness Evolution}

Harness-oriented work treats scaffolding as an evolving runtime object. Studies of harness composition, code as a runtime substrate, and readable scaffolds examine how that object is organized \cite{macedo,ning,rombaut,handbook}. Self-Harness, recursive-harness systems, and harness-benefit studies address revision and its relationship to later behavior \cite{selfharness,recursiveharness,linbenefit}. Monperrus connects coding-agent construction to bootstrapping \cite{monperrus}. Here, bootstrapping has a direct engineering meaning: a minimal working loop helps build the next layer of the harness, with each round drawing on capabilities developed earlier.

\section{Approach}

\subsection{Scope and Runtime}

We study one user, one agent, and one continuous session, with requests processed sequentially. Finishing each request before the next removes concurrency from the present account and keeps the control flow easy to inspect. The design relies only on context sent with the current call, so the harness resends any information intended to shape the next model response.

The runtime has three working areas. Model context contains material assembled for the current call. The tool environment comprises files, processes, services, and other resources available for actions to read or change. Persistent state carries history or memory beyond the current request. The harness transfers information among these areas. Incomplete responses and unsaved tool results remain local to the request in progress.

Consider a file-editing request. The user's instruction and selected history enter the model context. A proposed file change remains part of the model response until the harness accepts and dispatches the call. The harness turns the returned tool result into a request-local observation for the next model turn. Saving the request record then makes that observation available across requests. The same information changes role as it passes through the runtime, while model context, the tool environment, and persistent state retain separate responsibilities.

As a request-level contract, exposed tool names and argument shapes should match the dispatcher. The harness defines that tool set before the first model call and holds it fixed until the request ends. Configuration changes are deferred to later requests, preserving the meaning of calls already in progress.

Persistent state and the tool environment are updated independently. If a tool changes a file and saving the request record later fails, the external change may remain. The request record captures the interaction, while tool evidence supports claims about external effects. The harness records observed effects and labels uncertain execution or storage status instead of inferring success.

\subsection{Request Lifecycle}

Preparation reads the current input, selects relevant history or memory, loads descriptions of the available tools, and assembles the model context. When needed, housekeeping from an earlier request, such as saving an approved memory update, precedes context construction as a separate state operation.

A model response contains text, tool calls, or both. If the interface displays streamed text as progress, the harness still waits for the complete response before adding it to the request. It then decodes the response, checks the call identifiers, and validates every tool name and argument shape. Generation order determines the order of execution and observation return.

Invalid calls stop at validation and return a call-linked rejection observation, leaving the tool environment untouched. Valid calls enter the execution layer and yield a tool result: either a value or an expected error. The harness binds that result to the originating call, forms an observation, and includes it in the next model context.

Suppose one model response contains two calls: the first names an unknown tool, and the second a valid one. After checking both identifiers, the harness rejects the first call without execution and forms a rejection observation. It then runs the second call and forms a separate observation from the tool result. The next model turn receives both observations in their original order, preserving the distinction among model output, tool execution, and observation return.

The loop ends when the model returns no further calls or the harness reaches a local stop condition. The harness then assembles the request record and saves it. After a successful save, it sends the completion notification. Progress updates remain interface events rather than substitutes for the record. If the save fails, earlier tool effects may remain in the environment.

For later turns, the model context always includes the current instructions, input, and tool view. The selector draws optional history from the remaining context budget. Each retained observation identifies the call that produced its tool result. Labeled summaries condense older material and signal that details have been omitted.

As a design rule, shortened context should remain interpretable. Recent events often matter more than distant details, yet recency alone is insufficient. An observation separated from its call loses context, and a summary with no clear subject risks misleading the next turn. The selector therefore keeps the current request intact, preserves the link between each included observation and its action, and labels condensed material as a summary. The rule serves readability as well as memory.

\begin{figure*}[t]
  \centering
  \includegraphics[width=\textwidth]{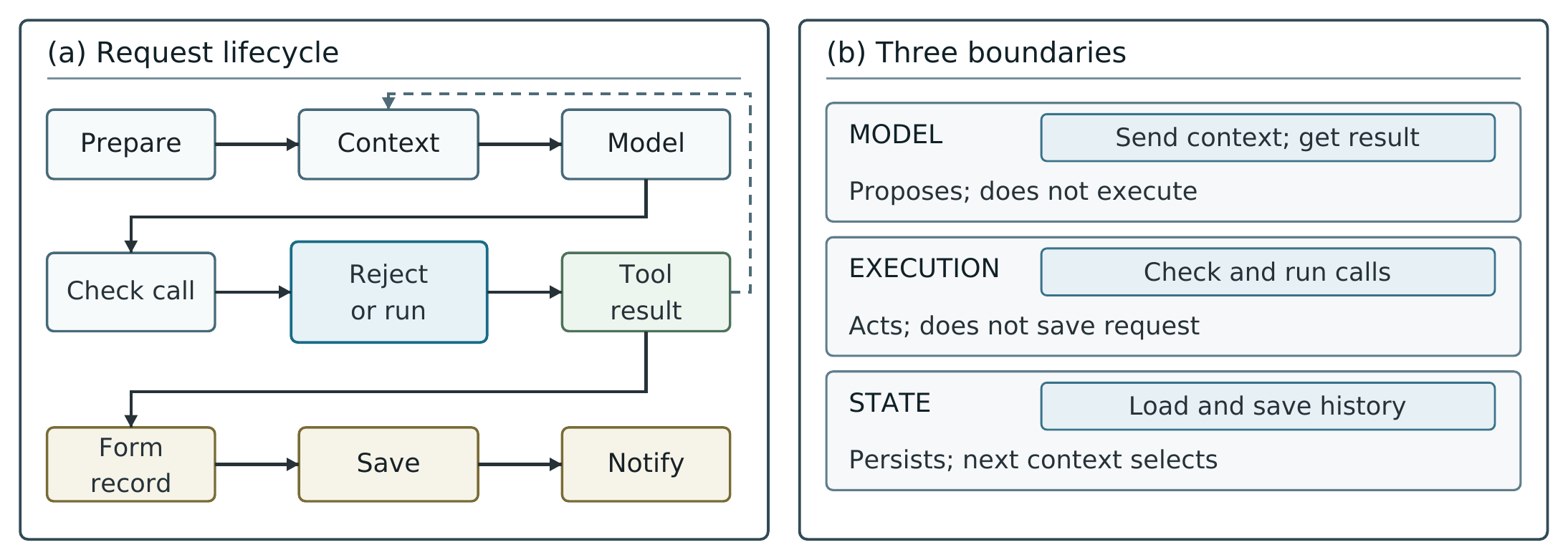}
  \caption{A compact harness viewed in two ways. Panel (a) follows one request from context preparation to saved completion. Panel (b) shows the model, execution, and state boundaries crossed during that request.}
  \label{fig:combined}
\end{figure*}

\subsection{Failures and Recovery}

Failure interpretation starts with the last completed stage. Under this design, rejection during validation produces an observation while leaving the environment unchanged. Once a tool effect occurs, a later request failure leaves that effect in place unless another action reverses it. A saved request record also stays in history when the completion notification never reaches the caller.

Storage and model visibility follow separate paths. Depending on the selector, material stored on disk may be omitted from a later context. For an uncertain execution status, the harness reports uncertainty in the current request instead of fabricating a successful observation. An uncertain save directs recovery toward storage inspection or a status check. Replay follows review because the original action may already have completed.

Failure analysis treats model output, environmental action, returned observations, request records, and user-facing completion as separately evidenced stages. This view makes partial traces easier to read and provides a clear recovery point.

A recovery procedure starts at the last confirmed stage. Before dispatch, correction or rejection is safe. After a confirmed tool effect, preserve the observation and handle any subsequent save failure as a state-layer problem. After a confirmed save, retry the completion notification without repeating the tool action. An uncertain external status calls for inspection or a tool-specific check before any replay.

\section{Harness}

\subsection{Mediation and Boundaries}

The request lifecycle orders events inside the harness, and the boundaries identify the resource crossed at each step. Model calls use the model service, tool calls enter the tool environment, and saves or loads access persistent state. Figure~\ref{fig:combined} presents the two views side by side and keeps the central loop visible within the larger runtime.

Execution handling turns an accepted model proposal into an action. The harness turns a tool result into a call-linked observation, and context assembly makes that observation available for later reasoning. Saving and later selecting a request record carries it forward. Most harness features extend one of these transitions.

\subsection{Boundary Contracts}

The model boundary handles protocol translation. The harness encodes context and tool descriptions for the service, then decodes the response into text and structured calls. Receipt confirms that the service returned a complete response. Later validation establishes call validity, and dispatch establishes execution.

Validation governs entry to the execution boundary. A valid call reaches the selected tool, while an invalid one returns as a rejection observation. The execution layer reports either a tool result---a value or an expected error---or an uncertain execution status. The harness records that report as a call-linked observation. Request persistence occurs later.

Persistence is handled at the state boundary, where the harness loads and saves history and memory. A successful save makes the request record available to later selectors. The selector then chooses the portion entering a new model context. Progress updates and the completion notification report status to the caller. Model, tool, and state facts come from the boundary operations that produced them.

The boundaries also localize debugging. Missing instructions direct attention to context construction. A valid proposal with no environmental effect leads to validation or dispatch. When a request record is absent from a later context, saving and selection become the relevant checks. This mapping narrows the search. It provides locations for evidence rather than a guarantee against failure.

\subsection{Consistency and Trust}

Several checks keep the three boundaries aligned. Call identifiers remain attached to their observations, and both calls and observations retain generation order. Names and arguments are validated before dispatch. Request records are saved before the completion notification is sent. Selection of older material preserves enough surrounding context to keep each call and observation understandable.

Useful checks include preserving call identifiers, returning observations in sequence, rejecting invalid names or arguments before execution, stopping a bounded inner loop, and saving a request record before the completion notification is sent. They exercise the actual control flow instead of merely restating the design.

When a compact implementation keeps a short history, trims context without preserving every related call--observation group, or leaves uncertain effects and delivery statuses untracked, its design documentation should expose those gaps. A scripted self-edit on a disposable copy demonstrates one bootstrap step: the running loop inspects, modifies, and tests mechanisms that affect later work. Repeating the step allows later runs to build on tested changes.

Persistence preserves content, not credibility. Potential weaknesses include omitted details in summaries, untrusted tool output, and outdated memory notes. Origin and status information help later decisions weigh that material. These measures support record keeping. Security isolation and content validation require separate mechanisms.

\section{Bootstrapping}

Bootstrapping begins with a working seed. A harness that reads a task, prepares model context, runs tools, returns observations, and retains useful state already provides the basic development loop. When editing and command tools expose the harness's own implementation, the same loop provides a way to work on the mechanisms that organize its behavior as well as on user projects.

The harness now plays two roles. During a request it coordinates the model, tools, observations, and state. Between runs, the same tools expose its context rules, tool interfaces, control flow, and state mechanisms for inspection and revision. The current loop provides a way to identify a missing ability, implement a focused change, run checks, and preserve the revision for later runs.

At the design level, revisions may give later runs richer context, a broader set of actions, or more useful state. Their practical benefit remains for separate empirical evaluation. Clearer context would support more demanding analysis, improved tools would widen the available operations, and richer observations or stronger state handling would help sustain longer tasks. These are pathways for gradual extension rather than measured gains reported here.

Human guidance fits this process. In a human-guided version, a person directs the work, reviews changes, and starts the next round from the accumulated implementation. The compact loop acts as a seed system, using its existing ability to interpret context, act, observe tool results, and retain state to help extend itself across successive runs.

\section{Conclusion}

Seen as a control layer, the harness supplies continuity across otherwise separate model calls. Its value lies in coordinating distinct transitions: proposals are checked before dispatch, observations return with their calls, and request records become reusable history through explicit saving and selection. This separation makes the runtime trace legible and gives recovery a concrete starting point. When the same loop also inspects and modifies the mechanisms that shape later runs, the compact implementation offers a practical seed for gradual extension.

\begingroup
\raggedright
\bibliographystyle{unsrt}
\bibliography{references}
\endgroup

\end{document}